\documentclass{article}

\usepackage{iclr2020_conference,times}

\pdfoutput=1

\usepackage[utf8]{inputenc} 
\usepackage[T1]{fontenc}    
\usepackage[hidelinks]{hyperref}
\usepackage{url}            
\usepackage{booktabs}       
\usepackage{amsfonts}       
\usepackage{nicefrac}       
\usepackage{microtype}      
\usepackage[title]{appendix}
\usepackage{tikz}
\usepackage{graphicx}
\usepackage{booktabs}
\usepackage{multirow}
\usepackage{enumitem}
\usepackage{xspace}
\usepackage{booktabs}
\usepackage{float}
\usepackage{color}
\usepackage{amsmath}
\usepackage{amssymb}
\usepackage{xcolor}
\usepackage[normalem]{ulem}
\usepackage{siunitx}
\usepackage{makecell}

\usepackage[font=small]{caption}
\usepackage{subcaption}
\usepackage{tabularx}

\usepackage{pgfplots,tikz}
\pgfplotsset{compat=1.9}

\usepackage{marvosym}

\bibpunct{(}{)}{;}{a}{}{,}

\newcommand{\vb}{\boldsymbol{b}}
\newcommand{\vx}{\boldsymbol{x}}
\newcommand{\vc}{\boldsymbol{c}}
\newcommand{\vh}{\boldsymbol{h}}
\newcommand{\vcprev}{\boldsymbol{c}_{\textit{prev}}}
\newcommand{\vhprev}{\boldsymbol{h}_{\textit{prev}}}
\newcommand{\mW}{\mathbf{W}}
\newcommand{\mQ}{\mathbf{Q}}
\newcommand{\mR}{\mathbf{R}}
\DeclareMathOperator*{\LSTM}{LSTM}
\DeclareMathOperator*{\mLSTM}{mLSTM}
\DeclareMathOperator*{\Mogrify}{Mogrify}

\newcommand{\ptb}{PTB\xspace}
\newcommand{\wikitexttwo}{Wikitext-2\xspace}
\newcommand{\enwik}{Enwik8\xspace}
\newcommand{\nlltoppl}[1]{%
  \pgfmathparse{exp(#1)}%
  \pgfmathprintnumber[fixed,zerofill,precision=1,assume math mode=true]{\pgfmathresult}}
\newcommand{\nlltopplbold}[1]{%
  \pgfmathparse{exp(#1)}%
  \textbf{\pgfmathprintnumber[fixed,zerofill,precision=1,assume math mode=true]{\pgfmathresult}}}
\newcommand{\nlltobpc}[1]{%
  \pgfmathparse{log2(exp(#1))}%
  \pgfmathprintnumber[fixed,zerofill,precision=3,assume math mode=true]{\pgfmathresult}}
\newcommand{\nlltobpcbold}[1]{%
  \pgfmathparse{log2(exp(#1))}%
  \textbf{\pgfmathprintnumber[fixed,zerofill,precision=3,assume math mode=true]{\pgfmathresult}}}

\usepackage{xargs} 
\usepackage[colorinlistoftodos,prependcaption,textsize=tiny]{todonotes}

\newcommand{\reducedstrut}{\vrule width 0pt height .9\ht\strutbox depth .9\dp\strutbox\relax}
\newcommand{\colorboxx}[2]{%
  \begingroup%
  \setlength{\fboxsep}{0pt}%
  \colorbox{#1}{\reducedstrut#2\/}%
  \endgroup
}
\newcommandx{\mgl}[2][1=]{\colorboxx{blue!30}{\tiny\,}\todo[color=blue!30,#1]{MG: #2}}
\newcommandx{\tk}[2][1=]{\colorboxx{brown!30}{\tiny\,}\todo[color=brown!30,#1]{TK: #2}}
\newcommandx{\pb}[2][1=]{\colorboxx{gray!30}{\tiny\,}\todo[color=gray!30,#1]{PB: #2}}
\newcommandx{\cd}[2][1=]{\colorboxx{cyan!30}{\tiny\,}\todo[color=cyan!30,#1]{CD: #2}}
\newcommandx{\jr}[2][1=]{\colorboxx{red!30}{\tiny\,}\todo[color=red!30,#1]{JR: #2}}

\pgfplotscreateplotcyclelist{acycle}{%
solid, every mark/.append style={solid, fill=gray}, mark=*\\%
dotted, every mark/.append style={solid, fill=gray}, mark=square*\\%
densely dotted, every mark/.append style={solid, fill=gray}, mark=otimes*\\%
loosely dotted, every mark/.append style={solid, fill=gray}, mark=triangle*\\%
dashed, every mark/.append style={solid, fill=gray},mark=diamond*\\%
loosely dashed, every mark/.append style={solid, fill=gray},mark=*\\%
densely dashed, every mark/.append style={solid, fill=gray},mark=square*\\%
dashdotted, every mark/.append style={solid, fill=gray},mark=otimes*\\%
dashdotdotted, every mark/.append style={solid},mark=star\\%
densely dashdotted,every mark/.append style={solid, fill=gray},mark=diamond*\\%
}

\allowdisplaybreaks

\title{Mogrifier LSTM}

\author{G\'abor Melis$^\dag$, Tom\'a\v{s} Ko\v{c}isk\'y$^\dag$, Phil Blunsom$^{\dag\ddag}$ \\
  {\tt \{melisgl,tkocisky,pblunsom\}@google.com}\\
  $^\dag$DeepMind, London, UK\\
  $^\ddag$University of Oxford
}

\iclrfinalcopy 
\begin{document}

\maketitle

\begin{abstract}
Many advances in Natural Language Processing have been based upon more
expressive models for how inputs interact with the context in which
they occur.
Recurrent networks, which have enjoyed a modicum of success, still
lack the generalization and systematicity ultimately required for
modelling language.
In this work, we propose an extension to the venerable Long Short-Term
Memory in the form of mutual gating of the current input and the
previous output.
This mechanism affords the modelling of a richer space of interactions
between inputs and their context.
Equivalently, our model can be viewed as making the transition
function given by the LSTM context-dependent.
Experiments demonstrate markedly improved generalization on language
modelling in the range of 3--4 perplexity points on Penn Treebank and
\wikitexttwo, and 0.01--0.05 bpc on four character-based datasets.
We establish a new state of the art on all datasets with the exception
of \enwik, where we close a large gap between the LSTM and Transformer
models.
\end{abstract}

\section{Introduction}

The domination of Natural Language Processing by neural models is
hampered only by their limited ability to generalize and questionable
sample complexity
\citep{belinkov2017synthetic,jia2017adversarial,iyyer2018adversarial,moosavi2017lexical,agrawal2016analyzing},
their poor grasp of grammar
\citep{linzen2016assessing,kuncoro2018lstms}, and their inability to
chunk input sequences into meaningful units \citep{wang2017sequence}.
While direct attacks on the latter are possible, in this paper, we
take a language-agnostic approach to improving Recurrent Neural
Networks (RNN, \cite{rumelhart1988learning}), which brought about many
advances in tasks such as language modelling, semantic parsing,
machine translation, with no shortage of non-NLP applications either
\citep{bakker2002reinforcement,mayer2008system}.
Many neural models are built from RNNs including the
sequence-to-sequence family \citep{sutskever2014sequence} and its
attention-based branch \citep{bahdanau2014neural}.
Thus, innovations in RNN architecture tend to have a trickle-down
effect from language modelling, where evaluation is often the easiest
and data the most readily available, to many other tasks, a trend
greatly strengthened by ULMFiT \citep{howard2018universal}, ELMo
\citep{peters2018deep} and BERT \citep{devlin2018bert}, which promote
language models from architectural blueprints to pretrained building
blocks.

To improve the generalization ability of language models, we propose
an extension to the LSTM \citep{hochreiter1997lstm}, where the LSTM's
input $\vx$ is gated conditioned on the output of the previous step
$\vhprev$.
Next, the gated input is used in a similar manner to gate the output
of the previous time step.
After a couple of rounds of this mutual gating, the last updated $\vx$
and $\vhprev$ are fed to an LSTM.
By introducing these additional of gating operations, in one sense,
our model joins the long list of recurrent architectures with gating
structures of varying complexity which followed the invention of Elman
Networks \citep{elman1990finding}.
Examples include the LSTM, the GRU \citep{chung2015gated}, and even
designs by Neural Architecture Search
\citep{DBLP:journals/corr/ZophL16}.

Intuitively, in the lowermost layer, the first gating step scales the
input embedding (itself a representation of the \emph{average} context
in which the token occurs) depending on the \emph{actual} context,
resulting in a contextualized representation of the input.
While intuitive, as Section\,\ref{sec:analysis} shows, this
interpretation cannot account for all the observed phenomena.

In a more encompassing view, our model can be seen as enriching the
mostly additive dynamics of recurrent transitions placing it in the
company of the Input Switched Affine Network \citep{foerster2017input}
with a separate transition matrix for each possible input, and the
Multiplicative RNN \citep{sutskever2011generating}, which factorizes
the three-way tensor of stacked transition matrices.
Also following this line of research are the Multiplicative
Integration LSTM \citep{wu2016multiplicative} and -- closest to our
model in the literature -- the Multiplicative LSTM
\citep{DBLP:journals/corr/KrauseLMR16}.
The results in Section\,\ref{sec:results} demonstrate the utility of
our approach, which consistently improves on the LSTM and establishes
a new state of the art on all but the largest dataset, \enwik, where
we match similarly sized transformer models.

\section{Model}



To allow for ease of subsequent extension, we present the standard
LSTM update \citep{DBLP:journals/corr/SakSB14} with input and state of
size $m$ and $n$ respectively as the following function:
\begin{gather*}
\LSTM \colon \mathbb{R}^m\times\mathbb{R}^n \times \mathbb{R}^n \to
\mathbb{R}^n \times \mathbb{R}^n \\
\LSTM(\vx, \vcprev, \vhprev) = (\vc, \vh).
\end{gather*}
The updated state $\vc$ and the output $\vh$ are computed as follows:
\begin{align*}
  \boldsymbol{f} &= \sigma(\mW^{fx} \vx + \mW^{fh} \vhprev + \vb^f) \\
  \boldsymbol{i} &= \sigma(\mW^{ix} \vx + \mW^{ih} \vhprev + \vb^i) \\
  \boldsymbol{j} &= \tanh(\mW^{jx}\vx + \mW^{jh} \vhprev + \vb^j) \\
  \boldsymbol{o} &= \sigma(\mW^{ox} \vx + \mW^{oh} \vhprev + \vb^o) \\
  \vc &= \boldsymbol{f} \odot \vcprev +
                \boldsymbol{i} \odot \boldsymbol{j} \\
  \boldsymbol{h} &= \boldsymbol{o} \odot \tanh(\vc),
\end{align*}
where $\sigma$ is the logistic sigmoid function, $\odot$ is the
elementwise product, $\mW^{**}$ and $b^{*}$ are weight matrices and
biases.

\begin{figure}[!t]\centering
  \includegraphics[scale=0.7,trim={1.8cm 7.5cm 8.3cm 2.5cm},clip]
                  {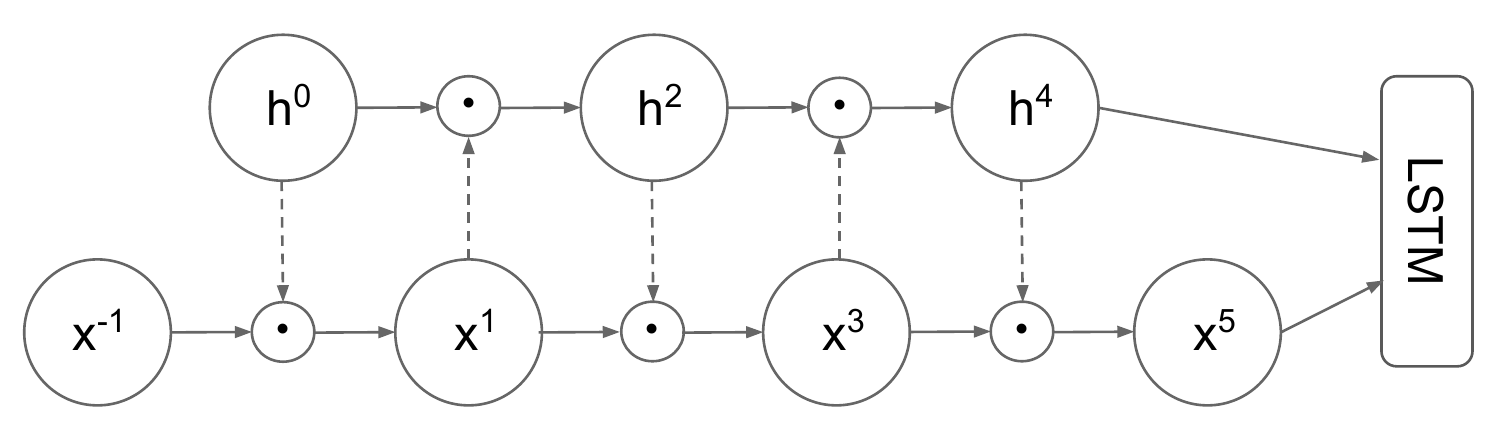}
  \caption{\small Mogrifier with 5 rounds of updates. The previous
    state $\vh^0=\vhprev$ is transformed linearly (dashed arrows), fed
    through a sigmoid and gates $\vx^{-1}=\vx$ in an elementwise
    manner producing $\vx^1$. Conversely, the linearly transformed
    $\vx^1$ gates $\vh^0$ and produces $\vh^2$. After a number of
    repetitions of this mutual gating cycle, the last values of
    $\vh^*$ and $\vx^*$ sequences are fed to an LSTM cell. The
    \emph{prev} subscript of $\vh$ is omitted to reduce clutter.}
  \label{fig:mogrifier}
\end{figure}
While the LSTM is typically presented as a solution to the vanishing
gradients problem, its gate $i$ can also be interpreted as scaling the
rows of weight matrices $\mW^{j*}$ (ignoring the non-linearity in
$j$).
In this sense, the LSTM nudges Elman Networks towards
context-dependent transitions and the extreme case of Input Switched
Affine Networks.
If we took another, larger step towards that extreme, we could end up
with Hypernetworks \citep{ha2016hypernetworks}.
Here, instead, we take a more cautious step, and equip the LSTM with
gates that scale the \emph{columns} of all its weight matrices
$\mW^{**}$ in a context-dependent manner.
The scaling of the matrices $\mW^{*x}$ (those that transform the cell
input) makes the input embeddings dependent on the cell state, while
the scaling of $\mW^{*h}$ does the reverse.

The Mogrifier\footnote{It's like a transmogrifier\footnotemark{}
  without the magic: it can only shrink or expand
  objects.}\footnotetext{Transmogrify (verb, 1650s): to completely
  alter the form of something in a surprising or magical manner.} LSTM
is an LSTM where two inputs $\vx$ and $\vhprev$ modulate one another
in an alternating fashion before the usual LSTM computation takes
place (see Fig.\,\ref{fig:mogrifier}).
That is, $\Mogrify(\vx, \vcprev, \vhprev) = \LSTM(\vx^{\uparrow},
\vcprev, \vhprev^{\uparrow})$ where the modulated inputs
$\vx^{\uparrow}$ and $\vhprev^{\uparrow}$ are defined as the highest
indexed $\vx^i$ and $\vhprev^i$, respectively, from the interleaved
sequences
\begin{align}
  \label{eq:Q}
  \vx^i &= 2\sigma(\mQ^{i}\vhprev^{i-1}) \odot \vx^{i-2}, &
  \text{for odd i} \in [1 \dots r]\\
  \label{eq:R}
  \vhprev^i &= 2\sigma(\mR^{i}\vx^{i-1}) \odot \vhprev^{i-2},
  & \text{for even i} \in [1 \dots r]
\end{align}
with $\vx^{-1} = \vx$ and $\vhprev^0 = \vhprev$.
The number of ``rounds'', $r \in \mathbb{N}$, is a hyperparameter;
$r=0$ recovers the LSTM.
Multiplication with the constant $2$ ensures that randomly initialized
$\mQ^{i}, \mR^{i}$ matrices result in transformations close to
identity.
To reduce the number of additional model parameters, we typically
factorize the $\mQ^{i}, \mR^{i}$ matrices as products of low-rank
matrices: $\mQ^i = \mQ^i_{\textrm{left}} \mQ^i_{\textrm{right}}$ with
$\mQ^i \in \mathbb{R}^{m \times n}, \mQ^i_{\textrm{left}} \in
\mathbb{R}^{m \times k}, \mQ^i_{\textrm{right}} \in \mathbb{R}^{k
  \times n}$, where $k<min(m,n)$ is the rank.

\section{Experiments}

\subsection{The Case for Small-Scale}

Before describing the details of the data, the experimental setup and
the results, we take a short detour to motivate work on smaller-scale
datasets.
A recurring theme in the history of sequence models is
that the problem of model design is intermingled with optimizability
and scalability.
Elman Networks are notoriously difficult to optimize, a property that
ultimately gave birth to the idea of the LSTM, but also to more recent
models such as the Unitary Evolution RNN \citep{arjovsky2016unitary}
and fixes like gradient clipping \citep{pascanu2013difficulty}.
Still, it is far from clear -- if we could optimize these models well
-- how different their biases would turn out to be.
The non-separability of model and optimization is fairly evident in
these cases.

Scalability, on the other hand, is often optimized for indirectly.
Given the limited ability of current models to generalize, we often
compensate by throwing more data at the problem.
To fit a larger dataset, model size must be increased.
Thus the best performing models are evaluated based on their
scalability\footnote{ Note that the focus on scalability is \emph{not}
  a problem per se. Indeed the unsupervised pretraining methods
  \citep{peters2018deep,devlin2018bert} take great advantage of this
  approach.}.
Today, scaling up still yields tangible gains on down-stream tasks,
and language modelling data is abundant.
However, we believe that simply scaling up will not solve the
generalization problem and better models will be needed.
Our hope is that by choosing small enough datasets, so that model size
is no longer the limiting factor, we get a number of practical
advantages:
\begin{itemize}[leftmargin=0.4cm]
\item[$\star$] Generalization ability will be more clearly reflected
  in evaluations even without domain adaptation.
\item[$\star$] Turnaround time in experiments will be reduced, and the
  freed up computational budget can be put to good use by controlling
  for nuisance factors.
\item[$\star$] The transient effects of changing hardware performance
  characteristics are somewhat lessened.
\end{itemize}

Thus, we develop, analyse and evaluate models primarily on small
datasets.
Evaluation on larger datasets is included to learn more about the
models' scaling behaviour and because of its relevance for
applications, but it is to be understood that these evaluations come
with much larger error bars and provide more limited guidance for
further research on better models.

\subsection{Datasets}

We compare models on both word and character-level language modelling
datasets.
The two word-level datasets we picked are the Penn Treebank (\ptb)
corpus by \citet{marcus1993building} with preprocessing from
\citet{mikolov2010recurrent} and \wikitexttwo by
\citet{DBLP:journals/corr/MerityXBS16}, which is about twice the size
of \ptb with a larger vocabulary and lighter preprocessing.
These datasets are definitely on the small side, but -- and
\emph{because} of this -- they are suitable for exploring different
model biases.
Their main shortcoming is the small vocabulary size, only in the tens
of thousands, which makes them inappropriate for exploring the
behaviour of the long tail.
For that, open vocabulary language modelling and byte pair encoding
\citep{sennrich2015neural} would be an obvious choice.
Still, our primary goal here is the comparison of the LSTM and
Mogrifier architectures, thus we instead opt for character-based
language modelling tasks, where vocabulary size is not an issue, the
long tail is not truncated, and there are no additional
hyperparameters as in byte pair encoding that make fair comparison
harder.
The first character-based corpus is \enwik from the Hutter Prize
dataset \citep{hutter2012human}.
Following common practice, we use the first 90 million characters for
training and the remaining 10 million evenly split between validation
and test.
The character-level task on the Mikolov preprocessed \ptb corpus
\citep{merity2018analysis} is unique in that it has the disadvantages
of closed vocabulary without the advantages of word-level modelling,
but we include it for comparison to previous work.
The final character-level dataset is the Multilingual Wikipedia Corpus
(MWC, \cite{kawakami2017learning}), from which we focus on the English
and Finnish language subdatasets in the single text, large setting.

\begin{table}
  \small
  \centering
  \caption{\small Word-level perplexities of near state-of-the-art
    models, our \textbf{LSTM} baseline and the \textbf{Mogrifier} on
    \ptb and \wikitexttwo. Models with Mixture of Softmaxes
    \citep{yang2017breaking} are denoted with \emph{MoS}, depth N with
    \emph{dN}. \emph{MC} stands for Monte-Carlo dropout evaluation.
    Previous state-of-the-art results in italics. Note the comfortable
    margin of 2.8--4.3 perplexity points the Mogrifier enjoys over the
    LSTM.}
  \label{tab:word-results}
  \begin{tabular}{@{}llrlrlr@{}}
    \toprule
    & & & \multicolumn{2}{c}{No Dyneval} & \multicolumn{2}{c}{Dyneval} \\
    \cmidrule(lr){4-5} \cmidrule(l){6-7}
    & & & Val. & Test & Val. & Test \\
    \midrule
    \parbox[t]{5mm}{\multirow{3}{*}{\rotatebox[origin=c]{90}{\parbox{2.4cm}{\centering PTB\\ EN}}}}
    & FRAGE (d3, MoS15) \citep{gong2018frage} & 22M
        & \emph{54.1} & \emph{52.4} & \emph{47.4} & \emph{46.5} \\
    & AWD-LSTM (d3, MoS15) \citep{yang2017breaking} & 22M
        & 56.5 & 54.4 & 48.3 & 47.7 \\
    & Transformer-XL \citep{dai2019transformer} & 24M
        & 56.7 & 54.5 & & \\
    & \textbf{LSTM} (d2) & 24M
        & \nlltoppl{4.02163} & \nlltoppl{3.99985}
        & \nlltoppl{3.88947} & \nlltoppl{3.88038} \\
    & \textbf{Mogrifier} (d2) & 24M
        & \nlltoppl{3.95346} & \nlltoppl{3.93124}
        & \nlltoppl{3.80963} & \nlltoppl{3.80708} \\
    & \textbf{LSTM} (d2, MC) & 24M
        & \nlltoppl{4.01559} & \nlltoppl{3.99003}
        & \nlltoppl{3.88338} & \nlltoppl{3.87987} \\
    & \textbf{Mogrifier} (d2, MC) & 24M
        & \nlltopplbold{3.93959} & \nlltopplbold{3.91415}
        & \nlltopplbold{3.80420} & \nlltopplbold{3.80321} \\
    \midrule
    \parbox[t]{5mm}{\multirow{3}{*}{\rotatebox[origin=c]{90}{\parbox{2.0cm}{\centering WT2\\ EN}}}}
    & FRAGE (d3, MoS15) \citep{gong2018frage} & 35M
        & \emph{60.3} & \emph{58.0} & \emph{40.8} & \emph{39.1} \\
    & AWD-LSTM (d3, MoS15) \citep{yang2017breaking} & 35M
        & 63.9 & 61.2 & 42.4 & 40.7 \\
    & \textbf{LSTM} (d2, MoS2) & 35M
        & \nlltoppl{4.13609} & \nlltoppl{4.09557}
        & \nlltoppl{3.76661} & \nlltoppl{3.72658} \\
    & \textbf{Mogrifier} (d2, MoS2) & 35M
        & \nlltoppl{4.07235} & \nlltoppl{4.03665}
        & \nlltoppl{3.70429} & \nlltoppl{3.66475} \\
    & \textbf{LSTM} (d2, MoS2, MC) & 35M
        & \nlltoppl{4.12526} & \nlltoppl{4.08426}
        & \nlltoppl{3.76540} & \nlltoppl{3.72385} \\
    & \textbf{Mogrifier} (d2, MoS2, MC) & 35M
        & \nlltopplbold{4.04771} & \nlltopplbold{4.00976}
        & \nlltopplbold{3.69399} & \nlltopplbold{3.65337} \\
    \bottomrule
  \end{tabular}
\end{table}

\subsection{Setup}

We tune hyperparameters following the experimental setup of
\cite{melis2018pushing} using a black-box hyperparameter tuner based
on batched Gaussian Process Bandits \citep{golovin2017google}.
For the LSTM, the tuned hyperparameters are the same:
\emph{input\_embedding\_ratio}, \emph{learning\_rate},
\emph{l2\_penalty}, \emph{input\_dropout},
\emph{inter\_layer\_dropout}, \emph{state\_dropout},
\emph{output\_dropout}.
For the Mogrifier, the number of rounds $r$ and the rank $k$ of the
low-rank approximation is also tuned (allowing for full rank, too).
For word-level tasks, BPTT \citep{werbos1990backpropagation} window
size is set to 70 and batch size to 64.
For character-level tasks, BPTT window size is set to 150 and batch
size to 128 except for \enwik where the window size is 500.
Input and output embeddings are tied for word-level tasks following
\cite{DBLP:journals/corr/InanKS16} and
\cite{DBLP:journals/corr/PressW16}.
Optimization is performed with Adam \citep{kingma2014adam} with
$\beta_1=0$, a setting that resembles RMSProp without momentum.
Gradients are clipped \citep{pascanu2013difficulty} to norm 10.
We switch to averaging weights similarly to
\cite{merity2017regularizing} after a certain number of checkpoints
with no improvement in validation cross-entropy or at 80\% of the
training time at the latest.
We found no benefit to using two-step finetuning.

Model evaluation is performed with the standard, deterministic dropout
approximation or Monte-Carlo averaging \citep{gal2016theoretically}
where explicitly noted (MC).
In standard dropout evaluation, dropout is turned off while in MC
dropout predictions are averaged over randomly sampled dropout masks
(200 in our experiments).
Optimal softmax temperature is determined on the validation set, and
in the MC case dropout rates are scaled \citep{melis2018pushing}.
Finally, we report results with and without dynamic evaluation
\citep{krause2017dynamic}.
Hyperparameters for dynamic evaluation are tuned using the same method
(see Appendix\,\ref{sec:hyperparameter-tuning-ranges} for details).

We make the code and the tuner output available at
\href{https://github.com/deepmind/lamb}{https://github.com/deepmind/lamb}.

\begin{table}
  \small
  \centering
  \caption{\small Bits per character on character-based datasets of
    near state-of-the-art models, our \textbf{LSTM} baseline and the
    \textbf{Mogrifier}. Previous state-of-the-art results in italics.
    Depth N is denoted with \emph{dN}. MC stands for Monte-Carlo
    dropout evaluation. Once again the Mogrifier strictly dominates
    the LSTM and sets a new state of the art on all but the \enwik
    dataset where with dynamic evaluation it closes the gap to the
    Transformer-XL of similar size ($\dag$ \cite{krause2019dynamic},
    $\ddag$ Ben Krause, personal communications, May 17, 2019). On
    most datasets, model size was set large enough for underfitting
    not to be an issue. This was very much not the case with \enwik,
    so we grouped models of similar sizes together for ease of
    comparison. Unfortunately, a couple of dynamic evaluation test
    runs diverged (NaN) on the test set and some were just too
    expensive to run (\enwik, MC).}
  \label{tab:character-results}
  \begin{tabular}{@{}llrllll@{}}
    \toprule
    & & & \multicolumn{2}{c}{No Dyneval} & \multicolumn{2}{c}{Dyneval} \\
    \cmidrule(lr){4-5} \cmidrule(lr){6-7}
    & & & Val. & \multicolumn{1}{r}{Test} & Val. & \multicolumn{1}{r}{Test} \\
    \midrule
    \parbox[t]{2mm}{\multirow{3}{*}{\rotatebox[origin=c]{90}
        {\parbox{2cm}{\centering PTB\\ EN}}}}
    & Trellis Networks \citep{bai2018trellis} & 13.4M & & \emph{1.159} & & \\
    & AWD-LSTM (d3) \citep{merity2017regularizing} & 13.8M & & 1.175 & & \\
    & \textbf{LSTM} (d2) & 24M
        & \nlltobpc{0.80641} & \nlltobpc{0.79259}
        & \nlltobpc{0.77349} & \nlltobpc{0.76448} \\
    & \textbf{Mogrifier} (d2) & 24M
        & \nlltobpc{0.79616} & \nlltobpc{0.78415}
        & \nlltobpc{0.76119} & \nlltobpc{0.75439} \\
    & \textbf{LSTM} (d2, MC) & 24M
        & \nlltobpc{0.80331} & \nlltobpc{0.78936}
        & \nlltobpc{0.77262} & \nlltobpc{0.76353} \\
    & \textbf{Mogrifier} (d2, MC) & 24M
        & \nlltobpcbold{0.78829} & \nlltobpcbold{0.77642}
        & \nlltobpcbold{0.75861} & \nlltobpcbold{0.75069} \\
    \midrule
    \multirow{3}{*}{\rotatebox[origin=c]{90}{\parbox{2cm}{\centering MWC\\ EN}}}
    & HCLM with Cache \citep{kawakami2017learning} & 8M
        & \emph{1.591} & \emph{1.538} & & \\
    & LSTM (d1) \citep{kawakami2017learning} & 8M
        & 1.793 & 1.736 & & \\
    & \textbf{LSTM} (d2) & 24M
        & \nlltobpc{0.93766} & \nlltobpc{0.92752}
        & \nlltobpc{0.85915} & \nlltobpc{0.84918} \\
    & \textbf{Mogrifier} (d2) & 24M
        & \nlltobpc{0.91464} & \nlltobpc{0.90439}
        & \nlltobpc{0.83346} & \nlltobpc{0.82386} \\
    & \textbf{LSTM} (d2, MC) & 24M
        & \nlltobpc{0.93336} & \nlltobpc{0.92356}
        & \nlltobpc{0.85783} & NaN \\
    & \textbf{Mogrifier} (d2, MC) & 24M
        & \nlltobpcbold{0.90977} & \nlltobpcbold{0.89980}
        & \nlltobpcbold{0.83193} & \nlltobpcbold{0.82302} \\
    \midrule
    \multirow{3}{*}{\rotatebox[origin=c]{90}{\parbox{2cm}{\centering MWC\\ FI}}}
    & HCLM with Cache \citep{kawakami2017learning} & 8M
        & \emph{1.754} & \emph{1.711} & & \\
    & LSTM (d1) \citep{kawakami2017learning} & 8M
        & 1.943 & 1.913 & & \\
    & \textbf{LSTM} (d2) & 24M
        & \nlltobpc{0.95833} & \nlltobpc{0.94772}
        & \nlltobpc{0.86577} & \nlltobpc{0.85727} \\
    & \textbf{Mogrifier} (d2) & 24M
        & \nlltobpc{0.92782} & \nlltobpc{0.91896}
        & \nlltobpc{0.83318} & \nlltobpcbold{0.82591} \\
    & \textbf{LSTM} (d2, MC) & 24M
        & \nlltobpc{0.95456} & \nlltobpc{0.94337}
        & \nlltobpc{0.86415} & \nlltobpc{0.85556} \\
    & \textbf{Mogrifier} (d2, MC) & 24M
        & \nlltobpcbold{0.91995} & \nlltobpcbold{0.91050}
        & \nlltobpcbold{0.83029} & NaN \\
    \midrule
    \parbox[t]{2mm}{\multirow{3}{*}{\rotatebox[origin=c]{90}
        {\parbox{4.8cm}{\centering \enwik\\ EN}}}}
    & Transformer-XL (d24) \citep{dai2019transformer} & 277M
        & & \textbf{0.993} & & \textbf{0.940}$\dag$ \\
    \cmidrule(l){2-7}
    & Transformer-XL (d18) \citep{dai2019transformer} & 88M
        & & 1.03 & & \\
    & \textbf{LSTM} (d4) & 96M
        & \nlltobpc{0.79343} & \nlltobpc{0.80062}
        & \nlltobpc{0.72191} & \nlltobpc{0.70699} \\
    & \textbf{Mogrifier} (d4) & 96M
        & \nlltobpc{0.76935} & \nlltobpc{0.77792}
        & \nlltobpc{0.69911} & \nlltobpc{0.68516} \\
    & \textbf{LSTM} (d4, MC) & 96M
        & \nlltobpc{0.78948} & \nlltobpc{0.79511}
        & & \\
    & \textbf{Mogrifier} (d4, MC) & 96M
        & \nlltobpc{0.76566} & \nlltobpc{0.77359}
        & & \\
    \cmidrule(l){2-7}
    & Transformer-XL (d12) \citep{dai2019transformer} & 41M
        & & 1.06 & & 1.01$\ddag$ \\
    & AWD-LSTM (d3) \citep{merity2017regularizing} & 47M
        & & 1.232 & & \\
    & mLSTM (d1) \citep{DBLP:journals/corr/KrauseLMR16} & 46M
        & & 1.24 & & 1.08 \\
    & \textbf{LSTM} (d4) & 48M
        & \nlltobpc{0.81915} & \nlltobpc{0.82841}
        & \nlltobpc{0.74416} & \nlltobpc{0.72877} \\
    & \textbf{Mogrifier} (d4) & 48M
        & \nlltobpc{0.78663} & \nlltobpc{0.79413}
        & \nlltobpc{0.71751} & \nlltobpc{0.70177} \\
    & \textbf{LSTM} (d4, MC) & 48M
        & \nlltobpc{0.81555} & \nlltobpc{0.82346}
        & & \\
    & \textbf{Mogrifier} (d4, MC) & 48M
        & \nlltobpc{0.78368} & \nlltobpc{0.79054}
        & & \\
    \bottomrule
  \end{tabular}
\end{table}

\subsection{Results}
\label{sec:results}

Table\,\ref{tab:word-results} lists our results on word-level
datasets.
On the \ptb and \wikitexttwo datasets, the Mogrifier has lower
perplexity than the LSTM by 3--4 perplexity points regardless of
whether or not dynamic evaluation \citep{krause2017dynamic} and
Monte-Carlo averaging are used.
On both datasets, the state of the art is held by the AWD LSTM
\citep{merity2017regularizing} extended with Mixture of Softmaxes
\citep{yang2017breaking} and FRAGE \citep{gong2018frage}.
The Mogrifier improves the state of the art without either of these
methods on \ptb, and without FRAGE on \wikitexttwo.

Table\,\ref{tab:character-results} lists the character-level modelling
results.
On all datasets, our baseline LSTM results are much better than those
previously reported for LSTMs, highlighting the issue of scalability
and experimental controls.
In some cases, these unexpectedly large gaps may be down to lack of
hyperparameter tuning as in the case of \cite{merity2017regularizing},
or in others, to using a BPTT window size (50) that is too small for
character-level modelling \citep{melis2017state} in order to fit the
model into memory.
The Mogrifier further improves on these baselines by a considerable
margin.
Even the smallest improvement of 0.012 bpc on the highly
idiosyncratic, character-based, Mikolov preprocessed \ptb task is
equivalent to gaining about 3 perplexity points on word-level \ptb.
MWC, which was built for open-vocabulary language modelling, is a much
better smaller-scale character-level dataset.
On the English and the Finnish corpora in MWC, the Mogrifier enjoys a
gap of 0.033-0.046 bpc.
Finally, on the \enwik dataset, the gap is 0.029-0.039 bpc in favour
of the Mogrifier.

Of particular note is the comparison to Transformer-XL
\citep{dai2019transformer}, a state-of-the-art model on larger
datasets such as Wikitext-103 and \enwik.
On \ptb, without dynamic evaluation, the Transformer-XL is on par with
our LSTM baseline which puts it about 3.5 perplexity points behind the
Mogrifier.
On \enwik, also without dynamic evaluation, the Transformer-XL has a
large, 0.09 bpc advantage at similar parameter budgets, but with
dynamic evaluation this gap disappears.
However, we did not test the Transformer-XL ourselves, so fair
comparison is not possible due to differing experimental setups and
the rather sparse result matrix for the Transformer-XL.

\section{Analysis}
\label{sec:analysis}

\subsection{Ablation Study}

The Mogrifier consistently outperformed the LSTM in our experiments.
The optimal settings were similar across all datasets, with $r \in
\{5,6\}$ and $k \in [40\dots90]$ (see
Appendix\,\ref{sec:hyperparameter-sensitivity} for a discussion of
hyperparameter sensitivity).
In this section, we explore the effect of these hyperparameters and
show that the proposed model is not unnecessarily complicated.
To save computation, we tune all models using a shortened schedule
with only 145 epochs instead of 964 and a truncated BPTT window size
of 35 on the word-level PTB dataset, and evaluate using the standard,
deterministic dropout approximation with a tuned softmax temperature.

\begin{figure}[!t]\centering
  \includegraphics[scale=0.7,trim={1.8cm 7.5cm 8.3cm 2.5cm},clip]
                  {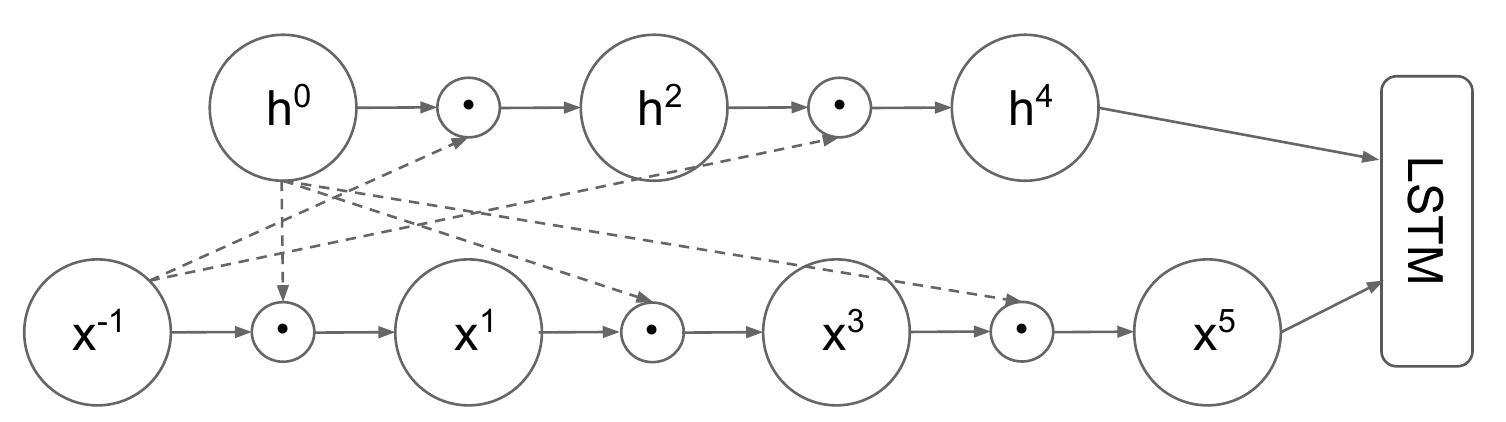}
  \caption{\small ``No-zigzag'' Mogrifier for the ablation study.
    Gating is always based on the original inputs.}
  \label{fig:mogrifier-no-zigzag}
\end{figure}

Fig.\,\ref{fig:ppl-vs-rounds} shows that the number of rounds $r$
greatly influences the results.
Second, we found the low-rank factorization of $\mQ^i$ and $\mR^i$ to
help a bit, but the full-rank variant is close behind which is what we
observed on other datasets, as well.
Finally, to verify that the alternating gating scheme is not overly
complicated, we condition \emph{all} newly introduced gates on the
original inputs $\vx$ and $\vhprev$ (see
Fig.\,\ref{fig:mogrifier-no-zigzag}).
That is, instead of Eq.\,\ref{eq:Q} and Eq.\,\ref{eq:R} the no-zigzag
updates are
\begin{align*}
  \vx^i &= 2\sigma(\mQ^{i}\vhprev) \odot \vx^{i-2} &
  \text{for odd i} \in [1 \dots r],\\
  \vhprev^i &= 2\sigma(\mR^{i}\vx) \odot \vhprev^{i-2}
  & \text{for even i} \in [1 \dots r].
\end{align*}
In our experiments, the no-zigzag variant underperformed the baseline
Mogrifier by a small but significant margin, and was on par with the
$r=2$ model in Fig.\,\ref{fig:ppl-vs-rounds} suggesting that the
Mogrifier's iterative refinement scheme does more than simply widen
the range of possible gating values of $\vx$ and $\vhprev$ to $(0,
2^{\lceil r/2 \rceil})$ and $(0, 2^{\lfloor r/2 \rfloor})$,
respectively.

\begin{figure}
\begin{minipage}{0.56\linewidth}
  \begin{tikzpicture}
    \begin{axis}[width=9.25cm, height=4.0cm,
        cycle list name=acycle]
	\addplot+ coordinates {
	  (0, 57.5) (1, 56.4) (2, 55.5)
          (3, 54.82) (4, 54.1) (5, 54.3) (6, 54.6)
	};
    \end{axis}
  \end{tikzpicture}
\captionof{figure}{Perplexity vs the rounds $r$ in the \ptb ablation
  study.}
\label{fig:ppl-vs-rounds}
\end{minipage}
\hfill
\begin{minipage}{0.36\linewidth}
\setlength\tabcolsep{28pt}
\captionof{table}{PTB ablation study
  validation perplexities with 24M parameters.}
\label{tab:ablation}
\begin{tabularx}{\textwidth}{@{}lr@{}}
  \toprule
  Mogrifier      & \nlltoppl{3.99163}  \\
  Full rank $Q^i,P^i$ & \nlltoppl{3.99947} \\
  No zigzag      & \nlltoppl{4.00647} \\
  LSTM           & \nlltoppl{4.051} \\
  mLSTM          & \nlltoppl{4.05750} \\
  \bottomrule
\end{tabularx}
\end{minipage}
\end{figure}

\subsection{Comparison to the mLSTM}
\label{seq:comparison-to-the-mlstm}

The Multiplicative LSTM \citep{DBLP:journals/corr/KrauseLMR16}, or
mLSTM for short, is closest to our model in the literature.
It is defined as $\mLSTM(\vx, \vcprev, \vhprev) = \LSTM(\vx, \vcprev,
\vhprev^m)$, where $\vhprev^m = (\mW^{mx} \vx) \odot (\mW^{mh} \vhprev)$.
In this formulation, the differences are readily apparent.
First, the mLSTM allows for multiplicative interaction between $\vx$
and $\vhprev$, but it only overrides $\vhprev$, while in the Mogrifier
the interaction is two-way, which -- as the ablation study showed --
is important.
Second, the mLSTM can change not only the magnitude but also the sign
of values in $\vhprev$, something with which we experimented in the
Mogrifier, but could not get to work.
Furthermore, in the definition of $\vhprev^m$, the unsquashed
linearities and their elementwise product make the mLSTM more
sensitive to initialization and unstable during optimization.

On the \enwik dataset, we greatly improved on the published results of
the mLSTM \citep{DBLP:journals/corr/KrauseLMR16}.
In fact, even our LSTM baseline outperformed the mLSTM by 0.03 bpc.
We also conducted experiments on \ptb based on our reimplementation of
the mLSTM following the same methodology as the ablation study and
found that the mLSTM did not improve on the LSTM (see
Table\,\ref{tab:ablation}).

\cite{DBLP:journals/corr/KrauseLMR16} posit and verify the recovery
hypothesis which says that having just suffered a large loss, the loss
on the next time step will be smaller on average for the mLSTM than
for the LSTM.
This was found not to be the case for the Mogrifier.
Neither did we observe a significant change in the gap between the
LSTM and the Mogrifier in the tied and untied embeddings settings,
which would be expected if recovery was affected by $\vx$ and
$\vhprev$ being in different domains.

\subsection{The Reverse Copy Task}

Our original motivation for the Mogrifier was to allow the context to
amplify salient and attenuate nuisance features in the input
embeddings.
We conduct a simple experiment to support this point of view.
Consider the reverse copy task where the network reads an input
sequence of tokens and a marker token after which it has to repeat the
input in reverse order.
In this simple sequence-to-sequence learning
\citep{sutskever2014sequence} setup, the reversal is intended to avoid
the minimal time lag problem \citep{hochreiter1997lstm}, which is not
our focus here.

The experimental setup is as follows.
For the training set, we generate \num{500000} examples by uniformly
sampling a given number of tokens from a vocabulary of size
\num{1000}.
The validation and test sets are constructed similarly, and contain
\num{10000} examples.
The model consists of an independent, unidirectional encoder and a
decoder, whose total number of parameters is \num{10} million.
The decoder is initialized from the last state of the encoder.
Since overfitting is not an issue here, no dropout is necessary, and
we only tune the learning rate, the l2 penalty, and the embedding size for
the LSTM.
For the Mogrifier, the number of rounds $r$ and the rank $k$ of the
low-rank approximation are also tuned.

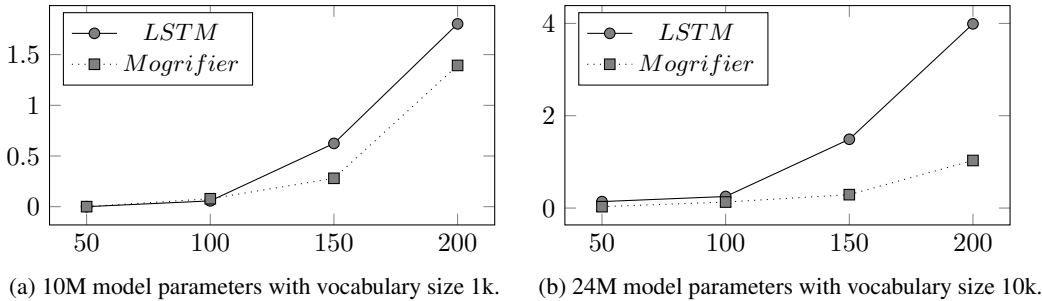
\begin{figure}[!t]
  \begin{subfigure}{0.49\linewidth}
  \begin{tikzpicture}
    \begin{axis}[width=7.5cm, height=4.5cm,
        legend pos = north west,
        legend style={font=\footnotesize},
        cycle list name=acycle]
	\addplot+ coordinates {
		(50, 0.00020) (100, 0.05715) (150, 0.62399) (200, 1.80348)
	};
	\addplot+ coordinates {
		(50, 0.00038) (100, 0.07821) (150, 0.27998) (200, 1.39342)
	};
    \legend{$LSTM$,$Mogrifier$}
    \end{axis}
  \end{tikzpicture}
  \caption{\small 10M model parameters with vocabulary size 1k.}
  \label{fig:copy-10m-results}
  \end{subfigure}
  \hfill
  \begin{subfigure}{0.49\linewidth}
  \begin{tikzpicture}
    \begin{axis}[width=7.5cm, height=4.5cm,
        legend pos = north west,
        legend style={font=\footnotesize},
        cycle list name=acycle]
	\addplot+ coordinates {
		(50, 0.14) (100, 0.25) (150, 1.49) (200, 3.99)
	};
	\addplot+ coordinates {
		(50, 0.03) (100, 0.13) (150, 0.29) (200, 1.03)
	};
    \legend{$LSTM$,$Mogrifier$}
    \end{axis}
  \end{tikzpicture}
  \caption{\small 24M model parameters with vocabulary size 10k.}
  \label{fig:copy-24m-results}
  \end{subfigure}
  \caption{\small Cross-entropy vs sequence length in the reverse copy
    task with i.i.d.\ tokens. Lower is better. The Mogrifier is better
    than the LSTM even in this synthetic task with no resemblance to
    natural language.}
\end{figure}

We compare the case where both the encoder and decoder are LSTMs to
where both are Mogrifiers.
Fig.\,\ref{fig:copy-10m-results} shows that, for sequences of length
50 and 100, both models can solve the task perfectly.
At higher lengths though, the Mogrifier has a considerable advantage.
Examining the best hyperparameter settings found, the embedding/hidden
sizes for the LSTM and Mogrifier are 498/787 vs 41/1054 at 150 steps,
and 493/790 vs 181/961 at 200 steps.
Clearly, the Mogrifier was able to work with a much smaller embedding
size than the LSTM, which is in line with our expectations for a model
with a more flexible interaction between the input and recurrent
state.
We also conducted experiments with a larger model and vocabulary size,
and found the effect even more pronounced (see
Fig.\,\ref{fig:copy-24m-results}).

\subsection{What the Mogrifier is not}

The results on the reverse copy task support our hypothesis that input
embeddings are enriched by the Mogrifier architecture, but that cannot
be the full explanation as the results of the ablation study indicate.
In the following, we consider a number of hypotheses about where the
advantage of the Mogrifier lies and the experiments that provide
evidence \emph{against} them.
\begin{itemize}[leftmargin=0.4cm]
\renewcommand\labelitemi{\scriptsize\Lightning}
\item \emph{Hypothesis: the benefit is in scaling $\vx$ and
  $\vhprev$.} We verified that data dependency is a crucial feature by
  adding a learnable scaling factor to the LSTM inputs. We observed no
  improvement. Also, at extremely low-rank (less than 5) settings
  where the amount of information in its gating is small, the
  Mogrifier loses its advantage.
\item \emph{Hypothesis: the benefit is in making optimization easier.}
  We performed experiments with different optimizers (SGD, RMSProp),
  with intra-layer batch normalization and layer normalization on the
  LSTM gates. While we cannot rule out an effect on optimization
  difficulty, in all of these experiments the gap between the LSTM and
  the Mogrifier was the same.
\item \emph{Hypothesis: exact tying of embeddings is too constraining,
  the benefit is in making this relationship less strict.} Experiments
  conducted with untied embeddings and character-based models
  demonstrate improvements of similar magnitude.
\item \emph{Hypothesis: the benefit is in the low-rank factorization
  of $\mQ^{i}, \mR^{i}$ implicitly imposing structure on the LSTM
  weight matrices.} We observed that the full-rank Mogrifier also
  performed better than the plain LSTM. We conducted additional
  experiments where the LSTM's gate matrices were factorized and
  observed no improvement.
\item \emph{Hypothesis: the benefit comes from better performance on
  rare words.} The observed advantage on character-based modelling is
  harder to explain based on frequency. Also, in the reverse copy
  experiments, a large number of tokens were sampled uniformly, so
  there were no rare words at all.
\item \emph{Hypothesis: the benefit is specific to the English
  language.} This is directly contradicted by the Finnish MWC and the
  reverse copy experiments.
\item \emph{Hypothesis: the benefit is in handling long-range
  dependencies better.} Experiments in the episodic setting (i.e.
  sentence-level language modelling) exhibited the same gap as the
  non-episodic ones.
\item \emph{Hypothesis: the scaling up of inputs saturates the
  downstream LSTM gates.} The idea here is that saturated gates may
  make states more stable over time. We observed the opposite: the
  means of the standard LSTM gates in the Mogrifier were very close
  between the two models, but their variance was smaller in the
  Mogrifier.
\end{itemize}

\section{Conclusions and Future Work}

We presented the Mogrifier LSTM, an extension to the LSTM, with
state-of-the-art results on several language modelling tasks.
Our original motivation for this work was that the context-free
representation of input tokens may be a bottleneck in language models
and by conditioning the input embedding on the recurrent state some
benefit was indeed derived.
While it may be part of the explanation, this interpretation clearly
does not account for the improvements brought by conditioning the
recurrent state on the input and especially the applicability to
character-level datasets.
Positioning our work on the Multiplicative RNN line of research offers
a more compelling perspective.

To give more credence to this interpretation, in the analysis we
highlighted a number of possible alternative explanations, and ruled
them all out to varying degrees.
In particular, the connection to the mLSTM is weaker than expected as
the Mogrifier does not exhibit improved recovery (see
Section\,\ref{seq:comparison-to-the-mlstm}), and on \ptb the mLSTM
works only as well as the LSTM.
At the same time, the evidence against easier optimization is weak,
and the Mogrifier establishing some kind of sharing between otherwise
independent LSTM weight matrices is a distinct possibility.

Finally, note that as shown by Fig.\,\ref{fig:mogrifier} and
Eq.\,\ref{eq:Q}-\ref{eq:R}, the Mogrifier is a series of preprocessing
steps composed with the LSTM function, but other architectures, such
as Mogrifier GRU or Mogrifier Elman Network are possible.
We also leave investigations into other forms of parameterization of
context-dependent transitions for future work.

\subsubsection*{Acknowledgments}

We would like to thank Ben Krause for the Transformer-XL dynamic
evaluation results, Laura Rimell, Aida Nematzadeh, Angeliki Lazaridou,
Karl Moritz Hermann, Daniel Fried for helping with experiments, Chris
Dyer, Sebastian Ruder and Jack Rae for their valuable feedback.

{
\small
\bibliography{paper}

\begin{thebibliography}{50}
\providecommand{\natexlab}[1]{#1}
\providecommand{\url}[1]{\texttt{#1}}
\expandafter\ifx\csname urlstyle\endcsname\relax
  \providecommand{\doi}[1]{doi: #1}\else
  \providecommand{\doi}{doi: \begingroup \urlstyle{rm}\Url}\fi

\bibitem[Agrawal et~al.(2016)Agrawal, Batra, and Parikh]{agrawal2016analyzing}
Aishwarya Agrawal, Dhruv Batra, and Devi Parikh.
\newblock Analyzing the behavior of visual question answering models.
\newblock \emph{arXiv preprint arXiv:1606.07356}, 2016.

\bibitem[Arjovsky et~al.(2016)Arjovsky, Shah, and Bengio]{arjovsky2016unitary}
Martin Arjovsky, Amar Shah, and Yoshua Bengio.
\newblock Unitary evolution recurrent neural networks.
\newblock In \emph{International Conference on Machine Learning}, pages
  1120--1128, 2016.

\bibitem[Bahdanau et~al.(2014)Bahdanau, Cho, and Bengio]{bahdanau2014neural}
Dzmitry Bahdanau, Kyunghyun Cho, and Yoshua Bengio.
\newblock Neural machine translation by jointly learning to align and
  translate.
\newblock \emph{arXiv preprint arXiv:1409.0473}, 2014.

\bibitem[Bai et~al.(2018)Bai, Kolter, and Koltun]{bai2018trellis}
Shaojie Bai, J~Zico Kolter, and Vladlen Koltun.
\newblock Trellis networks for sequence modeling.
\newblock \emph{arXiv preprint arXiv:1810.06682}, 2018.

\bibitem[Bakker(2002)]{bakker2002reinforcement}
Bram Bakker.
\newblock Reinforcement learning with long short-term memory.
\newblock In \emph{Advances in neural information processing systems}, pages
  1475--1482, 2002.

\bibitem[Belinkov and Bisk(2017)]{belinkov2017synthetic}
Yonatan Belinkov and Yonatan Bisk.
\newblock Synthetic and natural noise both break neural machine translation.
\newblock \emph{arXiv preprint arXiv:1711.02173}, 2017.

\bibitem[Chung et~al.(2015)Chung, Gulcehre, Cho, and Bengio]{chung2015gated}
Junyoung Chung, Caglar Gulcehre, Kyunghyun Cho, and Yoshua Bengio.
\newblock Gated feedback recurrent neural networks.
\newblock In \emph{International Conference on Machine Learning}, pages
  2067--2075, 2015.

\bibitem[Dai et~al.(2019)Dai, Yang, Yang, Cohen, Carbonell, Le, and
  Salakhutdinov]{dai2019transformer}
Zihang Dai, Zhilin Yang, Yiming Yang, William~W Cohen, Jaime Carbonell, Quoc~V
  Le, and Ruslan Salakhutdinov.
\newblock Transformer-xl: Attentive language models beyond a fixed-length
  context.
\newblock \emph{arXiv preprint arXiv:1901.02860}, 2019.

\bibitem[Devlin et~al.(2018)Devlin, Chang, Lee, and Toutanova]{devlin2018bert}
Jacob Devlin, Ming-Wei Chang, Kenton Lee, and Kristina Toutanova.
\newblock Bert: Pre-training of deep bidirectional transformers for language
  understanding.
\newblock \emph{arXiv preprint arXiv:1810.04805}, 2018.

\bibitem[Elman(1990)]{elman1990finding}
Jeffrey~L Elman.
\newblock Finding structure in time.
\newblock \emph{Cognitive science}, 14\penalty0 (2):\penalty0 179--211, 1990.

\bibitem[Foerster et~al.(2017)Foerster, Gilmer, Sohl-Dickstein, Chorowski, and
  Sussillo]{foerster2017input}
Jakob~N Foerster, Justin Gilmer, Jascha Sohl-Dickstein, Jan Chorowski, and
  David Sussillo.
\newblock Input switched affine networks: An rnn architecture designed for
  interpretability.
\newblock In \emph{Proceedings of the 34th International Conference on Machine
  Learning-Volume 70}, pages 1136--1145. JMLR. org, 2017.

\bibitem[Gal and Ghahramani(2016)]{gal2016theoretically}
Yarin Gal and Zoubin Ghahramani.
\newblock A theoretically grounded application of dropout in recurrent neural
  networks.
\newblock In \emph{Advances in Neural Information Processing Systems}, pages
  1019--1027, 2016.

\bibitem[Golovin et~al.(2017)Golovin, Solnik, Moitra, Kochanski, Karro, and
  Sculley]{golovin2017google}
Daniel Golovin, Benjamin Solnik, Subhodeep Moitra, Greg Kochanski, John Karro,
  and D~Sculley.
\newblock Google vizier: A service for black-box optimization.
\newblock In \emph{Proceedings of the 23rd ACM SIGKDD International Conference
  on Knowledge Discovery and Data Mining}, pages 1487--1495. ACM, 2017.

\bibitem[Gong et~al.(2018)Gong, He, Tan, Qin, Wang, and Liu]{gong2018frage}
Chengyue Gong, Di~He, Xu~Tan, Tao Qin, Liwei Wang, and Tie-Yan Liu.
\newblock Frage: frequency-agnostic word representation.
\newblock In \emph{Advances in Neural Information Processing Systems}, pages
  1334--1345, 2018.

\bibitem[Ha et~al.(2016)Ha, Dai, and Le]{ha2016hypernetworks}
David Ha, Andrew Dai, and Quoc~V Le.
\newblock Hypernetworks.
\newblock \emph{arXiv preprint arXiv:1609.09106}, 2016.

\bibitem[Hochreiter and Schmidhuber(1997)]{hochreiter1997lstm}
Sepp Hochreiter and J{\"u}rgen Schmidhuber.
\newblock Lstm can solve hard long time lag problems.
\newblock In \emph{Advances in neural information processing systems}, pages
  473--479, 1997.

\bibitem[Howard and Ruder(2018)]{howard2018universal}
Jeremy Howard and Sebastian Ruder.
\newblock Universal language model fine-tuning for text classification.
\newblock \emph{arXiv preprint arXiv:1801.06146}, 2018.

\bibitem[Hutter(2012)]{hutter2012human}
Marcus Hutter.
\newblock The human knowledge compression contest.
\newblock \emph{URL http://prize. hutter1. net}, 6, 2012.

\bibitem[Inan et~al.(2016)Inan, Khosravi, and
  Socher]{DBLP:journals/corr/InanKS16}
Hakan Inan, Khashayar Khosravi, and Richard Socher.
\newblock Tying word vectors and word classifiers: {A} loss framework for
  language modeling.
\newblock \emph{CoRR}, abs/1611.01462, 2016.
\newblock URL \url{http://arxiv.org/abs/1611.01462}.

\bibitem[Iyyer et~al.(2018)Iyyer, Wieting, Gimpel, and
  Zettlemoyer]{iyyer2018adversarial}
Mohit Iyyer, John Wieting, Kevin Gimpel, and Luke Zettlemoyer.
\newblock Adversarial example generation with syntactically controlled
  paraphrase networks.
\newblock \emph{arXiv preprint arXiv:1804.06059}, 2018.

\bibitem[Jia and Liang(2017)]{jia2017adversarial}
Robin Jia and Percy Liang.
\newblock Adversarial examples for evaluating reading comprehension systems.
\newblock \emph{arXiv preprint arXiv:1707.07328}, 2017.

\bibitem[Kawakami et~al.(2017)Kawakami, Dyer, and
  Blunsom]{kawakami2017learning}
Kazuya Kawakami, Chris Dyer, and Phil Blunsom.
\newblock Learning to create and reuse words in open-vocabulary neural language
  modeling.
\newblock \emph{arXiv preprint arXiv:1704.06986}, 2017.

\bibitem[Kingma and Ba(2014)]{kingma2014adam}
Diederik Kingma and Jimmy Ba.
\newblock Adam: A method for stochastic optimization.
\newblock \emph{arXiv preprint arXiv:1412.6980}, 2014.

\bibitem[Krause et~al.(2016)Krause, Lu, Murray, and
  Renals]{DBLP:journals/corr/KrauseLMR16}
Ben Krause, Liang Lu, Iain Murray, and Steve Renals.
\newblock Multiplicative {LSTM} for sequence modelling.
\newblock \emph{CoRR}, abs/1609.07959, 2016.
\newblock URL \url{http://arxiv.org/abs/1609.07959}.

\bibitem[Krause et~al.(2017)Krause, Kahembwe, Murray, and
  Renals]{krause2017dynamic}
Ben Krause, Emmanuel Kahembwe, Iain Murray, and Steve Renals.
\newblock Dynamic evaluation of neural sequence models.
\newblock \emph{arXiv preprint arXiv:1709.07432}, 2017.

\bibitem[Krause et~al.(2019)Krause, Kahembwe, Murray, and
  Renals]{krause2019dynamic}
Ben Krause, Emmanuel Kahembwe, Iain Murray, and Steve Renals.
\newblock Dynamic evaluation of transformer language models.
\newblock \emph{arXiv preprint arXiv:1904.08378}, 2019.

\bibitem[Kuncoro et~al.(2018)Kuncoro, Dyer, Hale, Yogatama, Clark, and
  Blunsom]{kuncoro2018lstms}
Adhiguna Kuncoro, Chris Dyer, John Hale, Dani Yogatama, Stephen Clark, and Phil
  Blunsom.
\newblock Lstms can learn syntax-sensitive dependencies well, but modeling
  structure makes them better.
\newblock In \emph{Proceedings of the 56th Annual Meeting of the Association
  for Computational Linguistics (Volume 1: Long Papers)}, pages 1426--1436,
  2018.

\bibitem[Linzen et~al.(2016)Linzen, Dupoux, and Goldberg]{linzen2016assessing}
Tal Linzen, Emmanuel Dupoux, and Yoav Goldberg.
\newblock Assessing the ability of lstms to learn syntax-sensitive
  dependencies.
\newblock \emph{Transactions of the Association for Computational Linguistics},
  4:\penalty0 521--535, 2016.

\bibitem[Marcus et~al.(1993)Marcus, Marcinkiewicz, and
  Santorini]{marcus1993building}
Mitchell~P Marcus, Mary~Ann Marcinkiewicz, and Beatrice Santorini.
\newblock Building a large annotated corpus of english: The {Penn} treebank.
\newblock \emph{Computational linguistics}, 19\penalty0 (2):\penalty0 313--330,
  1993.

\bibitem[Mayer et~al.(2008)Mayer, Gomez, Wierstra, Nagy, Knoll, and
  Schmidhuber]{mayer2008system}
Hermann Mayer, Faustino Gomez, Daan Wierstra, Istvan Nagy, Alois Knoll, and
  J{\"u}rgen Schmidhuber.
\newblock A system for robotic heart surgery that learns to tie knots using
  recurrent neural networks.
\newblock \emph{Advanced Robotics}, 22\penalty0 (13-14):\penalty0 1521--1537,
  2008.

\bibitem[Melis et~al.(2017)Melis, Dyer, and Blunsom]{melis2017state}
G{\'a}bor Melis, Chris Dyer, and Phil Blunsom.
\newblock On the state of the art of evaluation in neural language models.
\newblock \emph{arXiv preprint arXiv:1707.05589}, 2017.

\bibitem[Melis et~al.(2018)Melis, Blundell, Ko{\v{c}}isk{\`y}, Hermann, Dyer,
  and Blunsom]{melis2018pushing}
G{\'a}bor Melis, Charles Blundell, Tom{\'a}{\v{s}} Ko{\v{c}}isk{\`y},
  Karl~Moritz Hermann, Chris Dyer, and Phil Blunsom.
\newblock Pushing the bounds of dropout.
\newblock \emph{arXiv preprint arXiv:1805.09208}, 2018.

\bibitem[Merity et~al.(2016)Merity, Xiong, Bradbury, and
  Socher]{DBLP:journals/corr/MerityXBS16}
Stephen Merity, Caiming Xiong, James Bradbury, and Richard Socher.
\newblock Pointer sentinel mixture models.
\newblock \emph{CoRR}, abs/1609.07843, 2016.
\newblock URL \url{http://arxiv.org/abs/1609.07843}.

\bibitem[Merity et~al.(2017)Merity, Keskar, and Socher]{merity2017regularizing}
Stephen Merity, Nitish~Shirish Keskar, and Richard Socher.
\newblock Regularizing and optimizing lstm language models.
\newblock \emph{arXiv preprint arXiv:1708.02182}, 2017.

\bibitem[Merity et~al.(2018)Merity, Keskar, and Socher]{merity2018analysis}
Stephen Merity, Nitish~Shirish Keskar, and Richard Socher.
\newblock An analysis of neural language modeling at multiple scales.
\newblock \emph{arXiv preprint arXiv:1803.08240}, 2018.

\bibitem[Mikolov et~al.(2010)Mikolov, Karafi{\'a}t, Burget, Cernock{\`y}, and
  Khudanpur]{mikolov2010recurrent}
Tomas Mikolov, Martin Karafi{\'a}t, Lukas Burget, Jan Cernock{\`y}, and Sanjeev
  Khudanpur.
\newblock Recurrent neural network based language model.
\newblock In \emph{Interspeech}, volume~2, page~3, 2010.

\bibitem[Moosavi and Strube(2017)]{moosavi2017lexical}
Nafise~Sadat Moosavi and Michael Strube.
\newblock Lexical features in coreference resolution: To be used with caution.
\newblock \emph{arXiv preprint arXiv:1704.06779}, 2017.

\bibitem[Pascanu et~al.(2013)Pascanu, Mikolov, and
  Bengio]{pascanu2013difficulty}
Razvan Pascanu, Tomas Mikolov, and Yoshua Bengio.
\newblock On the difficulty of training recurrent neural networks.
\newblock In \emph{International conference on machine learning}, pages
  1310--1318, 2013.

\bibitem[Peters et~al.(2018)Peters, Neumann, Iyyer, Gardner, Clark, Lee, and
  Zettlemoyer]{peters2018deep}
Matthew~E Peters, Mark Neumann, Mohit Iyyer, Matt Gardner, Christopher Clark,
  Kenton Lee, and Luke Zettlemoyer.
\newblock Deep contextualized word representations.
\newblock \emph{arXiv preprint arXiv:1802.05365}, 2018.

\bibitem[Press and Wolf(2016)]{DBLP:journals/corr/PressW16}
Ofir Press and Lior Wolf.
\newblock Using the output embedding to improve language models.
\newblock \emph{CoRR}, abs/1608.05859, 2016.
\newblock URL \url{http://arxiv.org/abs/1608.05859}.

\bibitem[Rumelhart et~al.(1988)Rumelhart, Hinton, Williams,
  et~al.]{rumelhart1988learning}
David~E Rumelhart, Geoffrey~E Hinton, Ronald~J Williams, et~al.
\newblock Learning representations by back-propagating errors.
\newblock \emph{Cognitive modeling}, 5\penalty0 (3):\penalty0 1, 1988.

\bibitem[Sak et~al.(2014)Sak, Senior, and Beaufays]{DBLP:journals/corr/SakSB14}
Hasim Sak, Andrew~W. Senior, and Fran{\c{c}}oise Beaufays.
\newblock Long short-term memory based recurrent neural network architectures
  for large vocabulary speech recognition.
\newblock \emph{CoRR}, abs/1402.1128, 2014.
\newblock URL \url{http://arxiv.org/abs/1402.1128}.

\bibitem[Sennrich et~al.(2015)Sennrich, Haddow, and Birch]{sennrich2015neural}
Rico Sennrich, Barry Haddow, and Alexandra Birch.
\newblock Neural machine translation of rare words with subword units.
\newblock \emph{arXiv preprint arXiv:1508.07909}, 2015.

\bibitem[Sutskever et~al.(2011)Sutskever, Martens, and
  Hinton]{sutskever2011generating}
Ilya Sutskever, James Martens, and Geoffrey~E Hinton.
\newblock Generating text with recurrent neural networks.
\newblock In \emph{Proceedings of the 28th International Conference on Machine
  Learning (ICML-11)}, pages 1017--1024, 2011.

\bibitem[Sutskever et~al.(2014)Sutskever, Vinyals, and
  Le]{sutskever2014sequence}
Ilya Sutskever, Oriol Vinyals, and Quoc~V Le.
\newblock Sequence to sequence learning with neural networks.
\newblock In \emph{Advances in neural information processing systems}, pages
  3104--3112, 2014.

\bibitem[Wang et~al.(2017)Wang, Wang, Huang, Mohamed, Zhou, and
  Deng]{wang2017sequence}
Chong Wang, Yining Wang, Po-Sen Huang, Abdelrahman Mohamed, Dengyong Zhou, and
  Li~Deng.
\newblock Sequence modeling via segmentations.
\newblock In \emph{Proceedings of the 34th International Conference on Machine
  Learning-Volume 70}, pages 3674--3683. JMLR. org, 2017.

\bibitem[Werbos et~al.(1990)]{werbos1990backpropagation}
Paul~J Werbos et~al.
\newblock Backpropagation through time: what it does and how to do it.
\newblock \emph{Proceedings of the IEEE}, 78\penalty0 (10):\penalty0
  1550--1560, 1990.

\bibitem[Wu et~al.(2016)Wu, Zhang, Zhang, Bengio, and
  Salakhutdinov]{wu2016multiplicative}
Yuhuai Wu, Saizheng Zhang, Ying Zhang, Yoshua Bengio, and Ruslan~R
  Salakhutdinov.
\newblock On multiplicative integration with recurrent neural networks.
\newblock In \emph{Advances in neural information processing systems}, pages
  2856--2864, 2016.

\bibitem[Yang et~al.(2017)Yang, Dai, Salakhutdinov, and
  Cohen]{yang2017breaking}
Zhilin Yang, Zihang Dai, Ruslan Salakhutdinov, and William~W Cohen.
\newblock Breaking the softmax bottleneck: a high-rank rnn language model.
\newblock \emph{arXiv preprint arXiv:1711.03953}, 2017.

\bibitem[Zoph and Le(2016)]{DBLP:journals/corr/ZophL16}
Barret Zoph and Quoc~V. Le.
\newblock Neural architecture search with reinforcement learning.
\newblock \emph{CoRR}, abs/1611.01578, 2016.
\newblock URL \url{http://arxiv.org/abs/1611.01578}.

\end{thebibliography}
\bibliographystyle{plainnat}
}

\clearpage
\begin{appendices}

\section{Hyperparameter Tuning Ranges}
\label{sec:hyperparameter-tuning-ranges}

In all experiments, we tuned hyperparameters using Google Vizier
\citep{golovin2017google}.
The tuning ranges are listed in
Table\,\ref{tab:hyperparameter-tuning-ranges}.
Obviously, \emph{mogrifier\_rounds} and \emph{mogrifier\_rank} are
tuned only for the Mogrifier.
If $input\_embedding\_ratio \geqslant 1$, then the input/output
embedding sizes and the hidden sizes are set to equal and the linear
projection from the cell output into the output embeddings space is
omitted.
Similarly, $mogrifier\_rank \leqslant 0$ is taken to mean full rank
$\mQ^{*}$, $\mR^{*}$ without factorization.
Since \enwik is a much larger dataset, we don't tune
\emph{input\_embedding\_ratio} and specify tighter tuning ranges for
dropout based on preliminary experiments (see
Table\,\ref{tab:hyperparameter-tuning-ranges-enwik}).

Dynamic evaluation hyperparameters were tuned according to
Table\,\ref{tab:hyperparameter-tuning-ranges-dyneval}.
The highest possible value for \emph{max\_time\_steps}, the BPTT
window size, was 20 for word, and 50 for character-level tasks.
The batch size for estimating the mean squared gradients over the
training data was set to 1024, gradient clipping was turned off, and
the l2 penalty was set to zero.

\begin{table}[h!]
  \small
  \centering
  \caption{\small Hyperparameter tuning ranges for all tasks except
    \enwik.}
  \label{tab:hyperparameter-tuning-ranges}
  \begin{tabular}{@{}lrrrr@{}}
    \toprule
    & Low & High & Spacing \\
    learning\_rate & 0.001 & 0.004 & log \\
    input\_embedding\_ratio & 0.0 & 2.0 & \\
    l2\_penalty & 5e-6 & 1e-3 & log \\
    input\_dropout & 0.0 & 0.9 & \\
    inter\_layer\_dropout & 0.0 & 0.95 & \\
    state\_dropout & 0.0 & 0.8 \\
    output\_dropout & 0.0 & 0.95 \\
    mogrifier\_rounds ($r$) & 0 & 6 \\
    mogrifier\_rank ($k$) & -20 & 100 \\
    \bottomrule
  \end{tabular}
\end{table}

\begin{table}[h!]
  \small
  \centering
  \caption{\small Hyperparameter tuning ranges for \enwik.}
  \label{tab:hyperparameter-tuning-ranges-enwik}
  \begin{tabular}{@{}lrrrr@{}}
    \toprule
    & Low & High & Spacing \\
    learning\_rate & 0.001 & 0.004 & log \\
    l2\_penalty & 5e-6 & 1e-3 & log \\
    input\_dropout & 0.0 & 0.2 & \\
    inter\_layer\_dropout & 0.0 & 0.2 & \\
    state\_dropout & 0.0 & 0.25 \\
    output\_dropout & 0.0 & 0.25 \\
    mogrifier\_rounds ($r$) & 0 & 6 \\
    mogrifier\_rank ($k$) & -20 & 100 \\
    \bottomrule
  \end{tabular}
\end{table}

\begin{table}[h!]
  \small
  \centering
  \caption{\small Hyperparameter tuning ranges for dynamic evaluation.}
  \label{tab:hyperparameter-tuning-ranges-dyneval}
  \begin{tabular}{@{}lrrrr@{}}
    \toprule
    & Low & High & Spacing \\
    max\_time\_steps & 1 & 20/50 & \\
    dyneval\_learning\_rate & 1e-6 & 1e-3 & log \\
    dyneval\_decay\_rate & 1e-6 & 1e-2 & log \\
    dyneval\_epsilon & 1e-8 & 1e-2 & log \\
    \bottomrule
  \end{tabular}
\end{table}

\clearpage
\section{Hyperparameter Sensitivity}
\label{sec:hyperparameter-sensitivity}

\begin{figure}[!h]\centering
  \includegraphics[scale=0.5,trim={0cm 0.1cm 0.0cm 0cm},clip]
                  {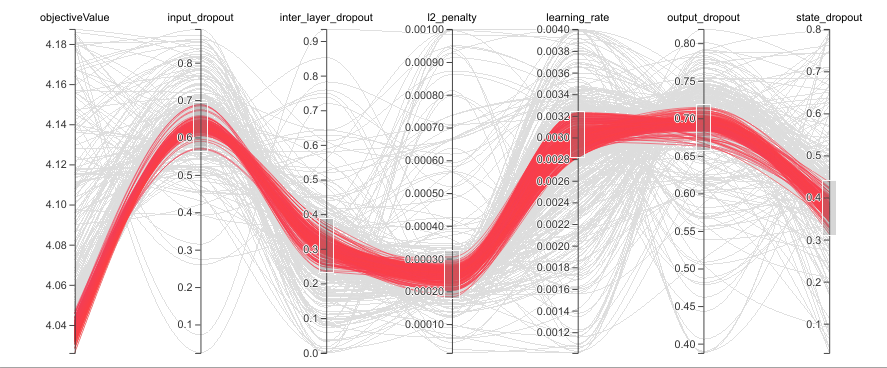}
  \caption{\small Average per-word validation cross-entropies for
    hyperparameter combinations in the neighbourhood of the best
    solution for a 2-layer LSTM with 24M weights on the Penn Treebank
    dataset.}
  \label{fig:lstm-sensitivity}
\end{figure}

\begin{figure}[!h]\centering
  \includegraphics[scale=0.5,trim={0cm 0.1cm 0.0cm 0cm},clip]
                  {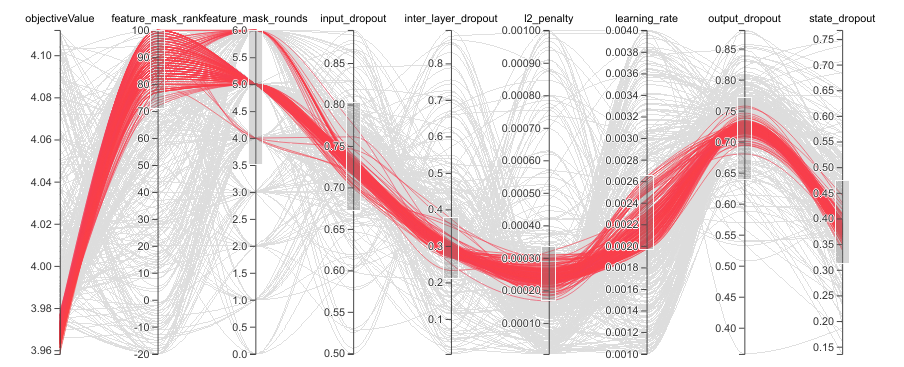}
  \caption{\small Average per-word validation cross-entropies for
    hyperparameter combinations in the neighbourhood of the best
    solution for a 2-layer Mogrifier LSTM with 24M weights on the Penn
    Treebank dataset. \emph{feature\_mask\_rank} and
    \emph{feature\_mask\_rounds} are aliases for
    \emph{mogrifier\_rank} and \emph{mogrifier\_rounds}}.
  \label{fig:fm-sensitivity}
\end{figure}

The parallel coordinate plots in Fig.\,\ref{fig:lstm-sensitivity} and
\ref{fig:fm-sensitivity}, give a rough idea about hyperparameter
sensitivity.
The red lines correspond to hyperparameter combinations closest to the
best solution found.
To find the closest combinations, we restricted the range for each
hyperparameter separately to about 15\% of its entire tuning range.

For both the LSTM and the Mogrifier, the results are at most 1.2
perplexity points off the best result, so our results are somewhat
insensitive to jitter in the hyperparameters.
Still, in this setup, grid search would require orders of magnitude
more trials to find comparable solutions.

On the other hand, the tuner does take advantage of the stochasticity
of training, and repeated runs with the same parameters may be give
slightly worse results.
To gauge the extent of this effect, on \ptb we estimated the standard
deviation in reruns of the LSTM with the best hyperparameters to be
about 0.2 perplexity points, but the mean was about 0.7 perplexity
points off the result produced with the weights saved in best tuning
run.

\end{appendices}

\end{document}